\documentclass[conference]{IEEEtran}
\usepackage{cite}
\usepackage{amsmath,amssymb,amsfonts}
\usepackage[ruled,vlined]{algorithm2e}
\usepackage{graphicx}
\usepackage{textcomp}
\usepackage{xcolor}
\usepackage{multirow}
\def\BibTeX{{\rm B\kern-.05em{\sc i\kern-.025em b}\kern-.08em
    T\kern-.1667em\lower.7ex\hbox{E}\kern-.125emX}}
\IEEEoverridecommandlockouts
\begin{document}

\title{Topology-Informed Visual Prompting For Vision Language Action Policies
}

\author{
Haoyang Wu$^{1}$, Abhinav Kumar$^{1}$, Dmitry Berenson$^{1}$%
\thanks{$^{1}$Robotics Department, University of Michigan, Ann Arbor, MI, USA.
{\tt\small \{haoyangw, abhin, dmitryb\}@umich.edu}}
}
\maketitle

\begin{abstract}
Vision-language-action (VLA) policies can struggle with manipulation tasks with complex obstacle geometries due to partial observability. 
These complex geometries can lead to similar visual observations or robot configurations requiring qualitatively different actions, a distinction that can be quantified using topological signatures.
While motion planners with full knowledge of environment geometries and object states can reason about these signatures in planning, this information is often not known at deployment.
To address this issue, we present a topology-guided visual-prompting framework that uses simulation-based planning to augment a nominal demonstration dataset and provides vision-based guidance at deployment.
Our method uses a Gauss-Linking-Integral topological signature representation to capture important topological properties of the environment. Using privileged geometry information from a simulation approximation of our environment, we augment a VLA fine-tuning dataset with trajectories that move the system to a demonstrated signature and, from the new configuration, resume task execution. A vision-language model (VLM) is fine-tuned on the same dataset to both predict signatures from live camera observations and predict end-effector waypoints, which are rendered as visual prompts on the observations to guide the VLA.
Across three simulated bimanual tasks and a real-world box pickup task, our method outperforms a VLA fine-tuned only on nominal demonstrations and a VLM-prompting baseline that can remove topology-relevant information from observations.
On hardware, it exceeds the strongest baseline by 40\% in task success. Project website: \texttt{https://topology-vla.github.io}.
\end{abstract}


\section{Introduction}
Vision-language-action (VLA) policies have shown great promise for learning generalizable policies from diverse data sources.
By combining large-scale Vision-Language Models (VLM) with action prediction, they aim to retain the generalizable intelligence of VLMs while adding action generation.
An important question, however, is whether VLAs operating under partial observability can reliably solve manipulation problems that are traditionally addressed assuming full observability.

Given ground truth information about environment geometries and object states, model-based planners can explicitly reason about geometric and topological constraints, the latter of which is the focus of this work.
Mitrano et al.~\cite{mitrano2024grasp} showed that for tasks where obstacle, robot, and environment geometries form closed loops, for example a robot grasping a cable threaded through part of the environment, as shown in Fig.~\ref{fig:threading_outcome}, explicitly reasoning about toplogical signatures can provide goal specifications and constraints for a planner.
Specifically, they use the h-signature \cite{bhattacharya2012topological}, a topological signature that describes how loops are linked.

While more diverse demonstration collection could improve VLA performance without requiring explicit topological reasoning, fine-tuned policies can still fail.
We frame our method as automating the augmentation of a nominal VLA fine-tuning dataset, enabling a large-scale automated version of data collection methods like DAgger \cite{ross2011reduction} through planning in simulation.
This allows us to take advantage of the capability of VLAs to operate under partial observability while still incorporating reasoning about topological signatures.

\begin{figure}[!t]
    \centering
    \includegraphics[width=\linewidth]{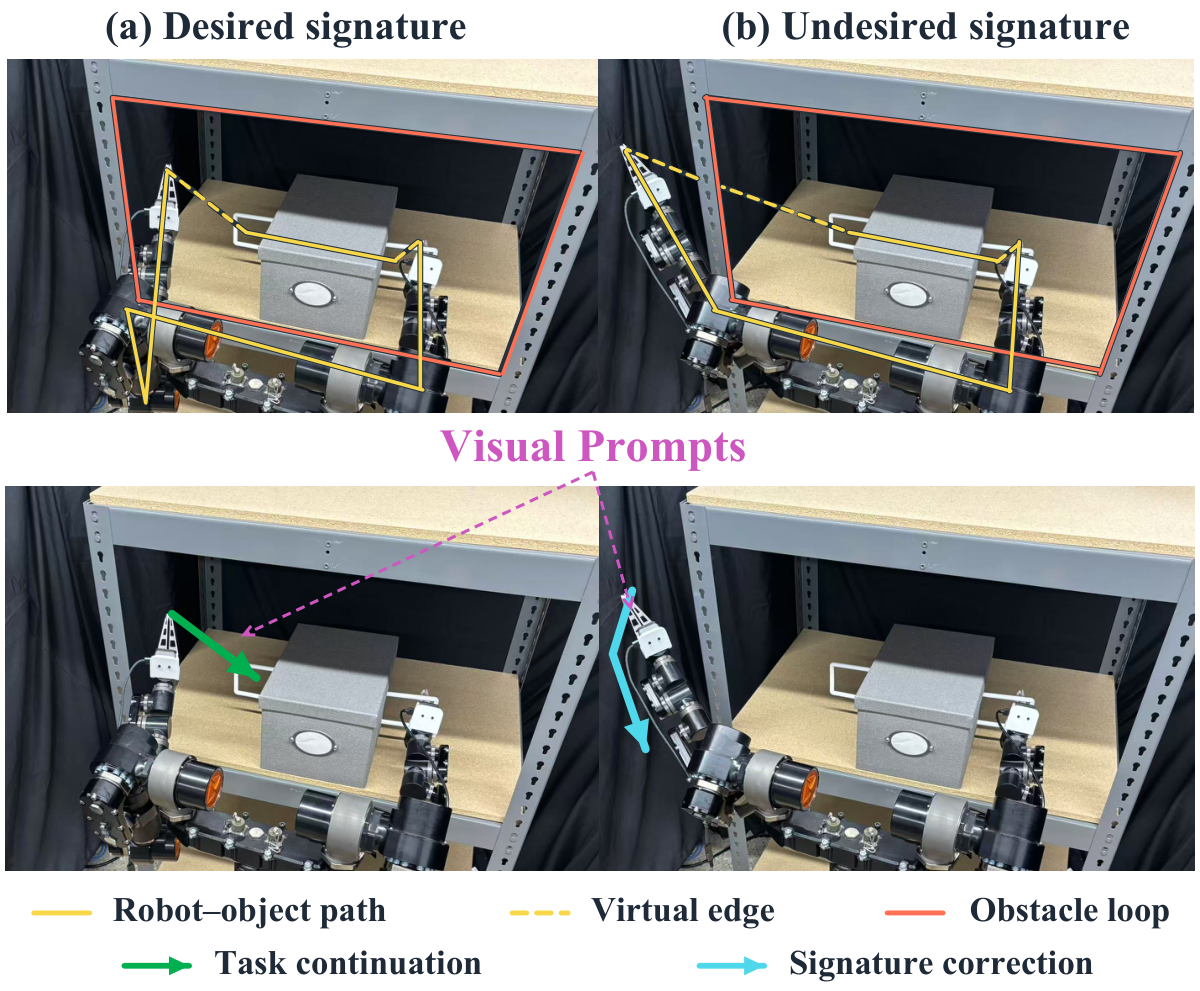}
    \caption{Different topological signatures require different motion directions for the same task goal. With the demonstrated signature, our green visual prompt guides the robot directly toward the handle. With an incorrect signature, our cyan prompt guides it around the rack post before approaching the handle.}
    \label{fig:topology_examples}
\end{figure}
While VLA models do include VLM backbones that could be fine-tuned to support topology reasoning, prior work~\cite{hancock2025actions,yang2025instructvla} has shown degradation in vision-language reasoning abilities of the VLM backbones after action training.
To account for this, we propose a framework that builds upon prior work in VLM prompting of VLAs~\cite{zhang2025peek}.

We fine-tune a VLM to identify topological signatures from vision inputs and, if the signature differs from the demonstrated signature for the task, output a visual prompt to guide the VLA to recover to the correct signature.
This visual prompt takes the form of an annotation drawn on the image that shows the needed end-effector motion to correct the signature.

The signature we use is a modification of the h-signature.
To ensure the signature is defined even when loops are open, for example, before the robot grasps an object, we adapt the h-signature formulation with \textit{virtual links} that close open loops, as shown in Fig.~\ref{fig:topology_examples}.
We refer to the signature computed from these virtually closed loops as the virtual h-signature.

Our contributions are:
\begin{itemize}
    \item A method for fine-tuning a pre-trained VLM to perform topological signature reasoning.
    \item A visual-prompting framework that augments VLA inputs with topology-informed guidance, generated by a VLM.
\end{itemize}

We evaluate our method on multiple topologically complex bimanual manipulation tasks in simulation, and a box pickup task in the real world.
In simulation and in the real world, our method improves success rate over a state-of-the-art VLM prompting baseline~\cite{zhang2025peek}, which can inadvertently remove toplogically-relevant information from VLA prompts. On hardware, our method outperforms the next best baseline by 40 percentage points.

\section{Related Works}
Our method expands upon prior work on augmenting VLAs through visual prompting, with those prompts being designed by a VLM.
To fine-tune the VLM to generate useful prompts, we apply prior work on topological motion planning.

\textbf{VLA Policies} Recent work has explored large, generalist robot policies trained across diverse tasks and embodiments \cite{brohan2022rt,o2024open,team2024octo}. 
    Building on this direction, vision-language-action (VLA) models leverage pretrained vision-language representations to connect semantic understanding with robot control \cite{brohan2023rt,kim2024openvla,black2024pi_0}. 
    In particular, $\pi_0$ combines a pretrained vision-language backbone with a flow-matching action expert to support general-purpose continuous robot control \cite{black2024pi_0}, while $\pi_{0.5}$ extends this through heterogeneous co-training across robot, semantic, and web data to improve generalization \cite{intelligence2025pi05visionlanguageactionmodelopenworld}. 
    
    GR00T N1 couples a vision-language module with a diffusion-transformer action module for generalist humanoid control \cite{nvidia2025gr00tn1openfoundation}, MolmoAct introduces depth-aware perception and explicit spatial trajectory reasoning before low-level action prediction \cite{lee2025molmoactactionreasoningmodels}, and MolmoBot trains generalist manipulation policies on large-scale simulated demonstrations for zero-shot transfer to real robots \cite{deshpande2026molmob0tlargescalesimulationenables}.
    
    However, these models still learn a direct mapping from visual observations, robot state, and language to actions. Nominal action supervision may not explicitly distinguish global relationships such as topology when locally similar inputs require different motions.
    In our experiments, we show that $\pi_0$ and $\pi_{0.5}$ models, even when fine-tuned on the tasks we consider, fail to learn generalizable topology reasoning but can be guided by our prompting framework.

\textbf{VLM-Guided Visual Prompting} Prior work uses image-space annotations to communicate spatial and motion information to vision-language models and robot policies.
These annotations include demonstrated trajectories and motion traces, candidate actions for VLM-based selection, and compact visual cues that guide manipulation policies \cite{gu2023rt,zheng2025tracevla,li2025hamster,zhang2025peek,nasiriany2024pivot}.
These works demonstrate that visual prompts can provide an effective intermediate interface between high-level reasoning and robot control, motivating our approach to distill the VLM's topological reasoning into visual guidance for the VLA.
However, they do not specifically consider topological signatures, leading to failures during task execution.
We compare against PEEK \cite{zhang2025peek}, a method that attempts to mask out irrelevant parts of observations and draw a path for an end-effector to follow.

\textbf{Topology-Guided Motion Planning} Prior work has explored motion planning in topologically complex environments, using topological concepts such as homotopy or homology classes to inform search \cite{bhattacharya2012topological,pokorny2016high,mitrano2024grasp}.
Other work uses topology to characterize robot--object interactions and manipulation states. 
Loop-based representations and quantities such as the Gauss linking integral have been used for grasping objects with holes, clasping and hooking, cooperative manipulation, and deformable-object manipulation \cite{pokorny2013grasping,stork2013topology,marzinotto2014cooperative,stork2013integrated,lui2013tangled,sudry2023hierarchical, bhattacharya2012topological}
These works assume full knowledge of obstacle geometries at deployment time.
In contrast, while we do assume access to a simulator in which we collect training data, our method does not require a high-fidelity simulation, allowing for discrepancies in obstacle geometries and layouts.
Using this approximate simulation, we generate training data that can be deployed in evaluation environments which may differ from the training data generation environments. 

\section{Preliminaries}
\label{sec:topology_representation}

The topological structures of interest are homotopy classes of task-relevant loops, defined in Mitrano et al.~\cite{mitrano2024grasp}.
Two loops belong to the same homotopy class if one can be continuously deformed into the other without intersecting an obstacle. For manipulation, we construct loops from task-relevant robot, object, and obstacle geometry. We compute this signature by extending the h-signature, defined as the Gauss linking number between a task-relevant robot--object loop and a task-relevant obstacle loop. The linking number is an integer that measures the signed winding of the two closed loops through each other~\cite{bhattacharya2012topological}. It is invariant to continuous deformation and can change only when loops pass through one another.

The physical robot and object paths in our tasks are open, so we extend the h-signature by connecting endpoints with virtual edges. As shown in Fig.~\ref{fig:topology_examples}, we first trace paths along the robot and manipulated object. Each robot path terminates at an end effector, which is paired with the endpoint of the object path at the corresponding grasp or contact site. These paths and endpoint pairings remain fixed across configurations of the task. At each configuration \(x_t\), a straight virtual edge connects each paired endpoint. For example, for a bimanual grasp through two handles, the two handles are the endpoints of the object path, and the virtual edges connect them to the corresponding end effectors. The physical paths and virtual edges together form the closed robot--object loop \(L_t\). We represent the task-relevant environmental structure by a separate closed obstacle loop \(O\). The virtual edges represent candidate straight-line end-effector paths to their target contact sites. Including these paths in \(L_t\) lets the signature indicate whether straight-line motion is compatible with the demonstrated virtual h-signature or whether the robot must first route around an obstacle.

For a configuration \(x_t\), its virtual h-signature $h^v=\operatorname{Lk}(L_t,O)$, where $\operatorname{Lk}$ is the Gauss Linking integral as used in \cite{bhattacharya2012topological}.
We define two configurations \(x\) and \(x'\) as having the same topological signature if and only if $h^v(x) =h^v(x')$. The same definition applies across tasks; only the physical paths, endpoint pairings, and obstacle loop used to construct \(L_t\) and \(O\) change with the task. This signature can therefore represent, for example, whether a cable passes through an aperture or whether a robot--object path is routed around an obstacle, independently of their poses and local geometry.
A virtual edge might pass through an obstacle, for example, if an end-effector is below the rack shown in Fig.~\ref{fig:topology_examples}.
Abstracting the scene into the $L_t$ and $O$ loops still allows us to compute $h^v$ and use it to guide the policy, for example moving the end-effector above the rack to achieve the demonstrated h-signature.

\section{Problem Statement}

We consider manipulation tasks whose success depends on achieving a specific \(h^v\). Let \(x_t \in \mathcal{X}\) denote the robot--object--environment configuration at time \(t\), and let \(h^v_t:=h^v(x_t)\) denote its \(h^v\)-signature at time $t$. We denote the demonstrated task signature by $h^{v \star}\in \mathbb{Z}$.
Because a mismatch \(h^v_t\neq h^{v \star}\) can prevent direct progress toward the goal, successful execution may require first transitioning the system into the demonstrated signature.
We assume that the task-relevant obstacle loop $O$ is provided by a demonstrator.
Given $O$ and a demonstration, we can infer the demonstrated signature \(h^{v \star}\).

We assume access to a simulator of our tasks, in which we collect data. The simulator need not exactly reproduce the geometry and appearance of the deployment environment, instead serving as an approximate model for computing $h^v$ values given $O$.
$O$ is only used to collect the simulation environment and is not assumed to be known in the real world.
From the simulator, we collect an offline dataset
\[
\mathcal{D}
=
\{\tau_i\}_{i=1}^{N},
\qquad
\tau_i
=
\{(o_t,s_t,h^v_t,a_t)\}_{t=0}^{T_i}.
\]
Here, \(o_t\) is the visual observation, \(s_t\) is the robot state, and \(a_t\) denotes the expert action chunk beginning at time \(t\). 
Each trajectory is also associated with a task instruction \(\ell\) and demonstrated h-signature \(h^{v \star}\).
The task-relevant obstacle geometry remains fixed within each rollout and is visible in \(o_t\).
Given \(\mathcal{D}\), our goal is to learn a manipulation policy that completes the desired task. 
To do this, we need a framework that detects whether the current $h^v$ matches the demonstrated $h^v$ and corrects a mismatch when needed.

At deployment, \(h^v_t\) is not observed. We assume that \(o_t\) contains sufficient visual information to infer $h^v_t$. Nevertheless, configurations with similar local geometry can have different $h^v$ and require qualitatively different manipulation strategies. Therefore, geometric proximity alone may not determine how the robot should progress toward the task goal. We evaluate a policy by its closed-loop task success rate, which measures whether it resolves $h^v$ mismatches when necessary and completes the manipulation task.

\section{Method}
\begin{figure*}[!t]
    \centering
    \includegraphics[width=\textwidth]{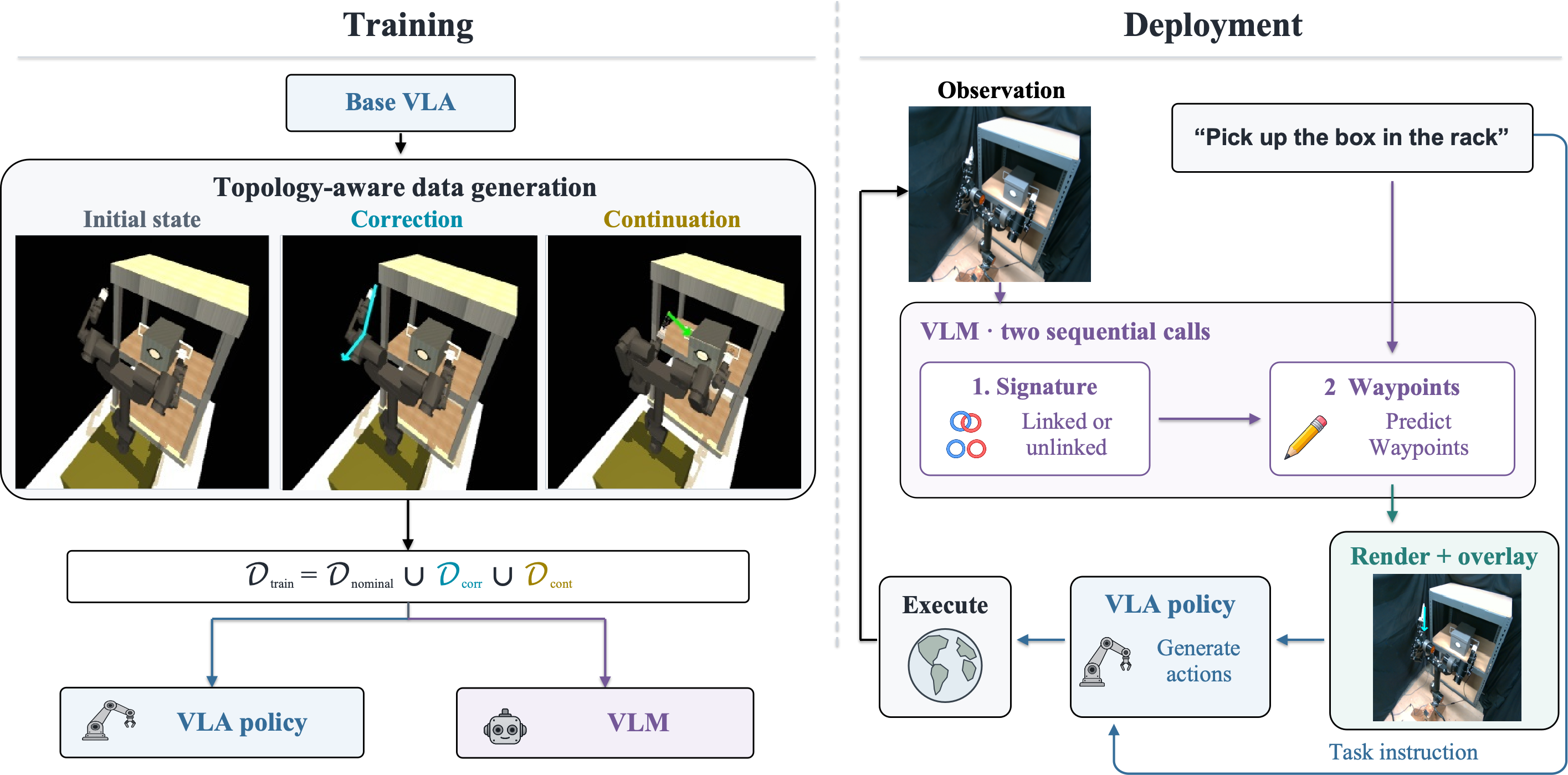}
    \caption{Overview of the training and deployment pipelines. During training, $h^v$-correction and task-continuation trajectories augment the nominal demonstrations used to supervise the VLA and VLM. At deployment, the VLM first predicts the current $h^v$ and then predicts image-space waypoints. {At each policy update, these waypoints are rendered on the current observation to guide the VLA's next action chunk.}}
    \label{fig:method_overview}
\end{figure*}

A standard VLA maps visual observations, robot state, and a language instruction directly to robot motion. In our setting, this mapping is ambiguous because similar visual inputs under the same instruction can correspond to different signatures and require different motions. By augmenting the observation input to the VLA, our method resolves this ambiguity.
\begin{figure}[!t]
    \centering
    \includegraphics[width=1.0\linewidth]{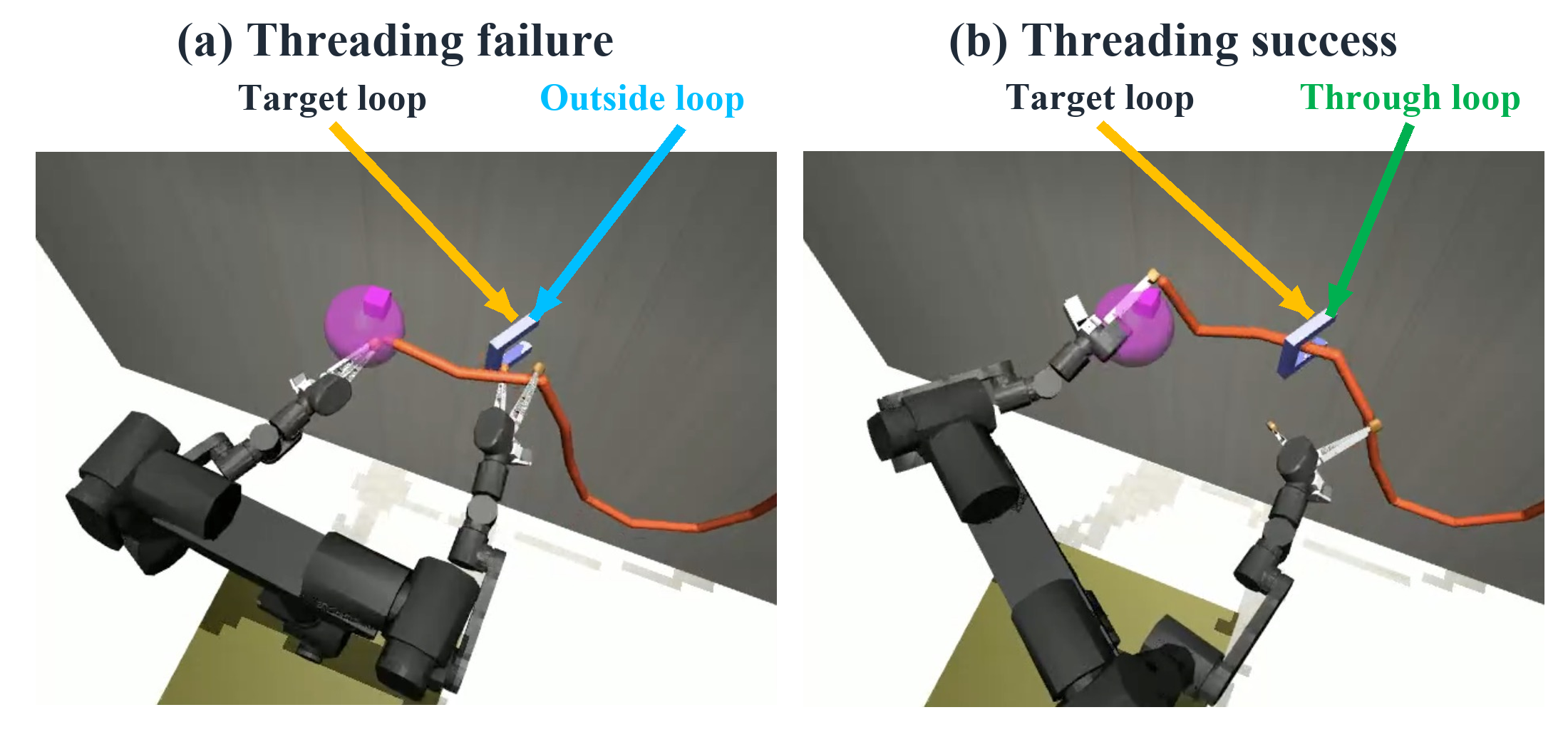}
    \caption{Successful and failed cable-threading rollouts. In (a), $\pi_{\mathrm{base}}$ moves the cable tip toward the goal while leaving it outside the loop. In (b), the cable passes through the target loop before reaching the purple goal.}
    \label{fig:threading_outcome}
\end{figure}
At a high level, we use a VLM to reason about the \(h^v\)-signature and, based on this reasoning, provide visual prompts that guide the VLA. Fig.~\ref{fig:method_overview} summarizes this architecture. We describe topology-aware demonstration generation in Sec.~\ref{sec:topology_aware_demonstrations}, construction of the visual prompts in Sec.~\ref{sec:visual_prompt_construction}, and VLM training and closed-loop guidance in Sec.~\ref{sec:vlm_guidance}.

\subsection{$h^v$-Aware Demonstration Generation}
\label{sec:topology_aware_demonstrations}

Our goal is to augment nominal demonstrations \(\mathcal{D}{\mathrm{nominal}}\) with a correction dataset \(\mathcal{D}{\mathrm{corr}}\)
which moves configurations to the demonstrated \(h^v\), and a continuation dataset \(\mathcal{D}_{\mathrm{cont}}\), which resumes task progress after correction, resulting in dataset
{
\begin{equation}
    \mathcal{D}_{\mathrm{train}}
    =
    \mathcal{D}_{\mathrm{nominal}}
    \cup
    \mathcal{D}_{\mathrm{corr}}
    \cup
    \mathcal{D}_{\mathrm{cont}}.
\end{equation}
}

We first train a base policy, $\pi_{\mathrm{base}}$, by fine-tuning a pretrained VLA on a nominal demonstration dataset $\mathcal{D}_{\mathrm{nominal}}$. 
These demonstrations cover successful task execution from randomized initial configurations, but they do not include explicit $h^v$ annotations or targeted recovery behaviors for correcting $h^v$ mismatches.
As a result, $\pi_{\mathrm{base}}$ captures the nominal manipulation skills required for task completion, while leaving $h^v$-specific recovery behaviors underrepresented.

Being in the incorrect $h^v$ could lead to a collision or block progress toward the goal. For example, as shown in Fig.~\ref{fig:threading_outcome}, a robot might move the cable tip toward the goal while leaving it outside the target loop. To augment the dataset with more diverse trajectories, we execute $\pi_{\mathrm{base}}$ in closed loop in simulation using the visual observation, robot state, and task instruction, and perturb states encountered along these rollouts.
Privileged $h^v$ labels from the simulator assign each perturbation resulting in a collision-free configuration to the appropriate recovery-data branch. The complete collection procedure is summarized in Algorithm~\ref{alg:topology_data_generation}.

\begin{algorithm}[t]
  \DontPrintSemicolon
  \caption{$h^v$-Aware Demonstration Generation}
  \label{alg:topology_data_generation}

  \KwIn{Base VLA $\pi_{\mathrm{base}}$, simulator $\mathcal{E}$,
  demonstrated h-signature $h^{v \star}$}
  \KwOut{Correction data $\mathcal{D}_{\mathrm{corr}}$ and
  continuation data $\mathcal{D}_{\mathrm{cont}}$}

  $\mathcal{D}_{\mathrm{corr}},\mathcal{D}_{\mathrm{cont}}
  \leftarrow\emptyset$\;

  \ForEach{task instance}{
      Initialize the task state $x_0$\;

      \While{the nominal rollout is active}{
          Execute $\pi_{\mathrm{base}}$ to obtain the next state $x_t$\;
          $\tilde{x}_t\leftarrow\textsc{PerturbState}(x_t)$\;

          \If{$\tilde{x}_t\neq\varnothing$}{
            Compute $h^v(\tilde{x}_t)$ using the simulator\;

            {
            \eIf{$h^v(\tilde{x}_t)\neq h^{v \star}$}{
                $\tau_{\mathrm{corr}}\leftarrow
                \textsc{CorrectSignature}(\tilde{x}_t,h^{v \star})$\;

                \If{$\tau_{\mathrm{corr}}=\varnothing$}{
                    \textbf{continue}\;
                }

                $\mathcal{D}_{\mathrm{corr}}\leftarrow
                \mathcal{D}_{\mathrm{corr}}
                \cup\{\tau_{\mathrm{corr}}\}$\;

                \mbox{$\tau_{\mathrm{cont}}\leftarrow\textsc{ContinueTask}(x^+,\pi_{\mathrm{base}},h^{v \star})$}\;
                \vspace{-0.6\baselineskip}
                \If{$\tau_{\mathrm{cont}}\neq\varnothing$}{
                    $\mathcal{D}_{\mathrm{cont}}\leftarrow
                    \mathcal{D}_{\mathrm{cont}}
                    \cup\{\tau_{\mathrm{cont}}\}$\;
                }
            }{
                \textbf{continue}\;
            }
            }
          }
      }
  }
  \vspace{-0.4\baselineskip}
  \end{algorithm}
At each collection step, \textsc{PerturbState} adds Gaussian noise to the current end-effector position, and a low-level controller moves the end effector toward the resulting target to produce \(\tilde{x}_t\). This motion is used only to construct the perturbed configuration and is not in the training dataset. Unreachable targets are discarded. For each perturbation resulting in a collision-free configuration, the simulator computes $h^v(\tilde{x}_t)$ and compares it with the demonstrated label $h^{v \star}$.

When $h^v(\tilde{x}_t) \neq h^{v \star}$, \textsc{CorrectSignature} searches for a motion that is collision-free with respect to the physical robot, object, and environment and that changes the label to the demonstrated class. We sample candidate Cartesian targets around the active end effector and use the workspace-goal-directed RRT formulation of Vande Weghe et al.~\cite{vande2007randomized} to plan a collision-free joint-space trajectory to each target. Each resulting trajectory is executed in simulation and retained only if its terminal configuration $x^+$ satisfies $h(x^+)=h^{v \star}$. When multiple valid corrections are found, we select the trajectory with the shortest end-effector path length.

After \textsc{CorrectSignature} changes a mismatched configuration to a corrected state \(x^+\), this state may lie outside the distribution covered by the nominal demonstrations. \textsc{ContinueTask} therefore executes \(\pi_{\mathrm{base}}\) from \(x^+\) and checks whether it reaches the next task subgoal while maintaining \(h(x)=h^{v \star}\). If the base VLA succeeds, \textsc{ContinueTask} returns \(\varnothing\), no continuation demonstration is added, and data collection proceeds to the next state of the nominal rollout. If the base VLA fails, a low-level end-effector IK controller generates a fallback continuation trajectory. We add only this fallback trajectory to \(\mathcal{D}_{\mathrm{cont}}\), provided that it reaches the subgoal while maintaining \(h(x)=h^{v \star}\). Otherwise, no continuation demonstration is added. If perturbation or correction fails, no demonstration is added for that candidate.

We train the final policy, \(\pi_{\mathrm{VLA}}\), as a separate fine-tuning run initialized from the original pretrained VLA rather than from \(\pi_{\mathrm{base}}\). All trajectories in \(\mathcal{D}_{\mathrm{train}}\) are pooled without balancing or reweighting the three data sources.
We describe how these end-effector trajectories are converted into visual prompts for VLM training below.

\begin{figure*}[!t]
\centering
\begin{minipage}[t]{0.32\textwidth}
\centering
{\small{\textbf{(a)} Rope Pulling}}\par\smallskip
\includegraphics[width=\linewidth]{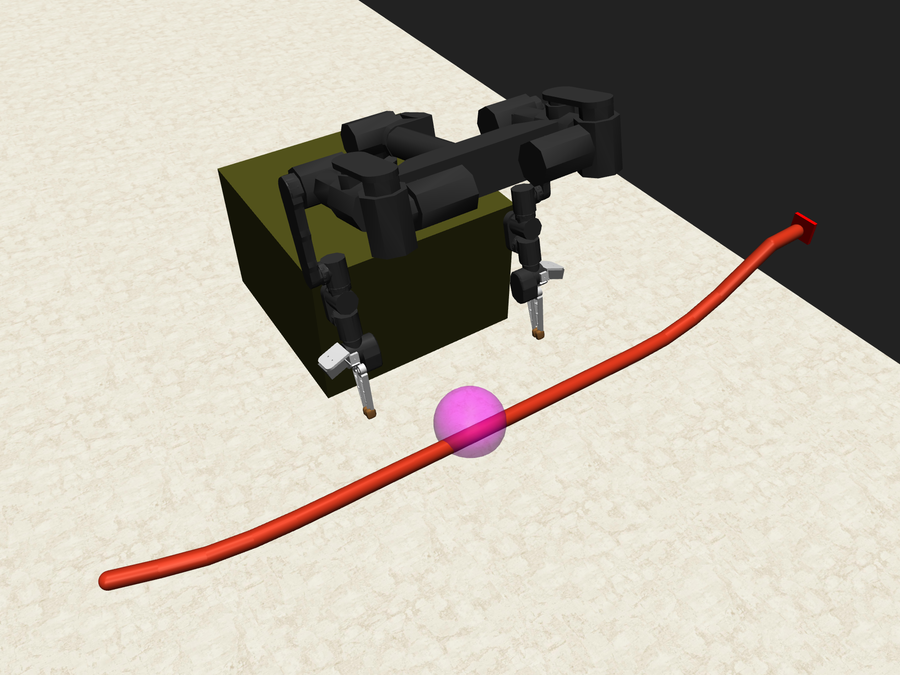}
\end{minipage}\hfill
\begin{minipage}[t]{0.32\textwidth}
\centering
{\small{\textbf{(b)} Cable Threading}}\par\smallskip
\includegraphics[width=\linewidth]{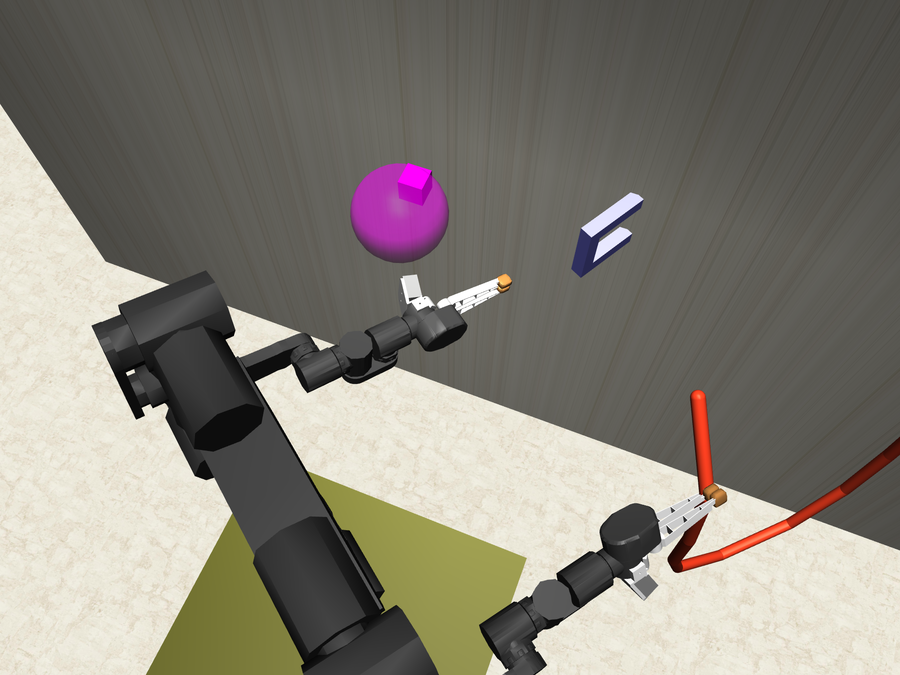}
\end{minipage}\hfill
\begin{minipage}[t]{0.32\textwidth}
\centering
{\small{\textbf{(c)} Box Pickup}}\par\smallskip
\includegraphics[width=\linewidth]{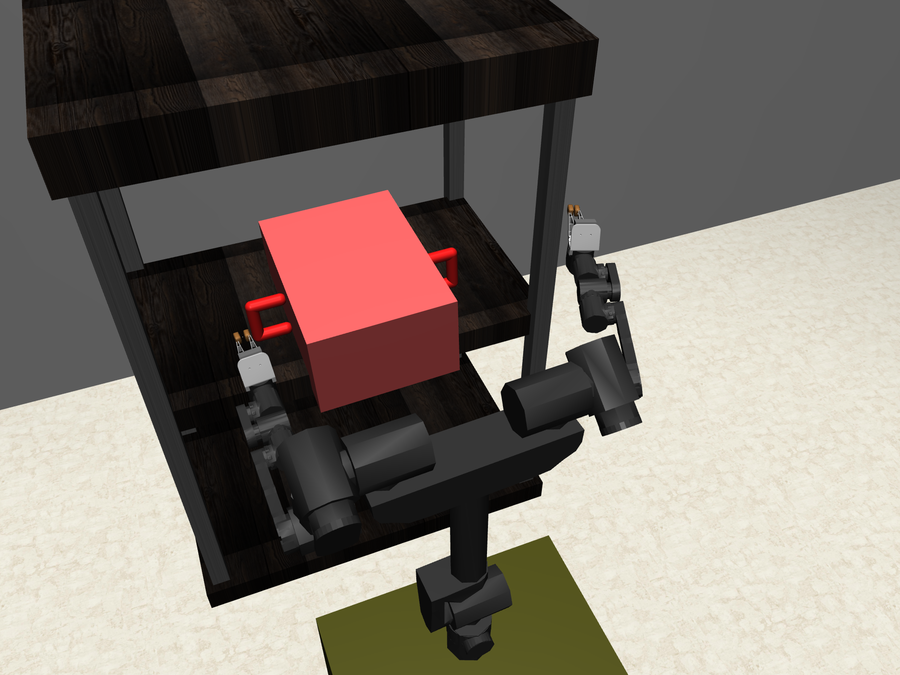}
\end{minipage}
\caption{Representative initial simulation observations for (a) Rope Pulling, (b) Cable Threading, and (c) Box Pickup.}
\label{fig:simulation_tasks}
\end{figure*}

\subsection{Visual Prompt Construction}
\label{sec:visual_prompt_construction}

We separate topology reasoning from action prediction because jointly adapting a VLM for reasoning and robot actions can degrade its pretrained reasoning capabilities~\cite{hancock2025actions,yang2025instructvla}. Following prior work on trajectory annotations~\cite{gu2023rt,zheng2025tracevla}, we communicate \(h^v\) guidance through image-space drawings without modifying the VLA architecture.

We convert the end-effector trajectories generated in Sec.~\ref{sec:topology_aware_demonstrations} into image-space drawings that are overlaid on the VLA's visual observations. At each timestep, we project the desired end-effector motion onto the image from the scene camera. We denote the ordered image-space waypoint coordinates by \(W_t\). A deterministic renderer connects these waypoints in temporal order and adds an arrowhead to indicate the direction of motion. {We overlay this drawing on the visual observation \(o_t\) to obtain the prompted observation \(\widetilde{o}_t\).}

We construct visual prompts for both $h^v$-correction and task-continuation trajectories. $h^v$-correction trajectories are rendered as arrows depicting the motion required to reach $h^{v*}$. 
These prompts consist of full trajectories and are capable of representing non-linear end-effector motion, for example a curved path around the post of a rack, by using multiple waypoints.
Task-continuation trajectories are rendered as arrows in a different color from the topology-correction arrows, pointing toward the next task subgoal, such as a grasp location. Nominal task demonstrations are left unmodified. Each prompted observation \(\widetilde{o}_t\) is paired with the expert action chunk \(a_t\), allowing the VLA to learn how image-space guidance maps to continuous control. The projected waypoint coordinates \(W_t\) also serve as supervision targets for the VLM described in the next subsection. Privileged quantities, including \(h^v_t\) and three-dimensional planning targets, are not provided to the policy.

\subsection{Closed-Loop VLM Guidance}
\label{sec:vlm_guidance}

We train the VLA using drawings constructed from privileged $h^v$ labels and expert trajectories, which are unavailable at deployment. We therefore train a single VLM to recognize the current $h^v$ from camera observations and then predict the image-space waypoint coordinates. We separate these tasks because $h^v$ labels are easier to obtain than expert trajectories.

We train the VLM with two types of supervised examples: 1) $h^v$-prediction examples pair a visual observation \(o_t\) with its simulator-provided label \(h^v_t\); and 2) Waypoint-prediction examples pair \((o_t,\ell,h^v_t{,h^{v \star}})\) with the ordered image-space coordinates \(W_t\) constructed in Sec.~\ref{sec:visual_prompt_construction}. {Here, \(h^v_t\) provides the ground-truth current signature during training, and \(h^{v \star}\) specifies the demonstrated signature.} For nominal training states, the waypoint target is an empty sequence.

At deployment, we query the same VLM twice. We denote the $h^v$ and waypoint calls by \(f_{h^v}\) and \(f_{\mathrm{way}}\). The first call predicts the virtual h-signature from the visual observation,
\[
\widehat{h}^v_t
=
f_{h^v}(o_t).
\]
Because the ground-truth {current} label is unavailable at deployment, the second call uses \(\widehat{h}^v_t\) with the visual observation, task instruction, and known demonstrated signature \(h^{v \star}\),
\[
\widehat{W}_t
=
f_{\mathrm{way}}(o_t,\ell,\widehat{h}^v_t{,h^{v \star}}).
\]
The output is an ordered sequence of image-space waypoint coordinates. We render these coordinates as an arrow using the same visual conventions employed during VLA training, then overlay the resulting transparent layer on \(o_t\) to obtain \(\widetilde{o}_t\). An empty waypoint sequence leaves the observation unchanged. The VLA then maps the prompted observation, robot state, and original task instruction to a continuous action chunk,
\[
\widehat{a}_{t:t+H_a-1}
=
\pi_{\mathrm{VLA}}
\left(
\widetilde{o}_t,
s_t,
\ell
\right),
\]
where \(H_a\) is the action-chunk horizon. After executing each predicted chunk, we acquire a new observation and repeat both VLM calls. This closed-loop process updates the drawing along task progress.

\section{Results}

Our experiments address four questions. Does our method improve task performance across different \(h^{v\star}\) values and VLA backbones, remain effective when only one \(h^v\) is possible, and transfer to real hardware? We evaluate Cable Threading and Box Pickup as tasks with multiple possible $h^v$ values, Rope Pulling as a task with only a single possible $h^v$, and Box Pickup on hardware to assess real-world feasibility.

\subsection{Evaluation Tasks}

We consider three tasks in simulation: Rope Pulling, Cable Threading, and Box Pickup.
For all tasks, we design simulation environments and task execution oracles to autonomously generate data for VLA and VLM fine-tuning.

\subsubsection{Rope Pulling}

Fig.~\ref{fig:simulation_tasks}(a) shows a representative initial observation. We evaluate each method over 50 trials with randomized rope, goal, and robot configurations. Because the rope tip begins beyond the robot's reachable workspace, the robot alternates arms to grasp and pull it into the goal region. Since $h^v_t=0$ throughout, no signature correction is required. This task tests whether our pipeline remains compatible with general learning without degrading nominal performance.

\subsubsection{Cable Threading}

{Fig.~\ref{fig:simulation_tasks}(b) shows a representative initial observation. We evaluate each method over 100 simulation trials.}
Cable Threading requires a transition from $h^v_0=0$, where the cable has not passed through the target loop, to $h^{v\star}=1$. Its free tip is initially held by the right gripper on one side of the loop, with the left gripper near the opposite side. We randomize the cable configuration, robot configuration, and target-loop position across trials. Because moving the tip around the loop may bring it near the goal without achieving the required signature, the oracle passes the tip through the loop with the right gripper, secures it with the left gripper, and releases it with the right gripper. The left arm then pulls the tip toward the goal. A trial succeeds when the cable remains through the loop and its tip reaches the purple goal region without being dropped.

\subsubsection{Box Pickup}

{Fig.~\ref{fig:simulation_tasks}(c) shows an initial observation in simulation. We evaluate each method over 100 simulation trials and 30 real-world trials.}
Box Pickup requires correcting $h^v_0=1$ to $h^{v\star}=0$ before lifting the box. The box begins within a rack, with both grippers outside its two handle openings. We randomize the rack size, box position, and robot configuration across trials. In trials with $h^v_0=1$, the right arm lies between the rack posts, so moving directly toward its handle would cause a collision. The robot must first route the arm around the posts to reach $h^v_t=0$, then approach the handles, pass each gripper through its corresponding opening, and lift the box. A trial succeeds when the box clears the rack while maintaining the demonstrated signature and both grasps.

\begin{table}[!t]
\caption{Task success rates (\%) in simulation.
\\ Base Policy refers to the backbone fine-tuned on $\mathcal{D}_{\mathrm{nominal}}$.}
\label{tab:simulation_results}
\centering
\small
\setlength{\tabcolsep}{3.5pt}
\begin{tabular}{llcc}
\hline
& & \multicolumn{2}{c}{VLA backbone} \\
Task & Method & \(\pi_0\)~\cite{black2024pi_0} & \(\pi_{0.5}\)~\cite{intelligence2025pi05visionlanguageactionmodelopenworld} \\
\hline
\multirow{3}{*}{Rope Pulling} & Base Policy & \textbf{70} & 52 \\
& PEEK~\cite{zhang2025peek} & 54 & 50 \\
& Ours & 66 & 64 \\
\hline
\multirow{3}{*}{Cable Threading} & Base Policy & 49 & 47 \\
& PEEK~\cite{zhang2025peek} & 13 & 20 \\
& Ours & 71 & \textbf{77} \\
\hline
\multirow{3}{*}{Box Pickup} & Base Policy & 31 & 45 \\
& PEEK~\cite{zhang2025peek} & 57 & 66 \\
& Ours & \textbf{84} & 82 \\
\hline
\end{tabular}
\end{table}

\subsection{Implementation Details}


We use \(\pi_0\)~\cite{black2024pi_0} and \(\pi_{0.5}\)~\cite{intelligence2025pi05visionlanguageactionmodelopenworld} as VLA backbones and Qwen3-VL-8B-Instruct~\cite{bai2025qwen3} for both VLM calls. For each backbone, the base policy, PEEK, and our method use the same VLA training configuration.

We fine-tune each VLA for 20,000 steps with a global batch size of 32, a learning rate of \(5\times10^{-5}\), and 1,000 warm-up steps. We use a 10-step action horizon, \(224\times224\) images, bfloat16 precision, and LoRA~\cite{hu2021lora} ranks of 16 for the 2B vision-language backbone and 32 for the 300M action expert. Training uses four NVIDIA A40 GPUs. During execution, all 10 actions in a predicted chunk are completed before the next VLM and VLA update.

We fine-tune the VLM for $h^v$ and waypoint prediction for one epoch using LoRA~\cite{hu2021lora} with rank 8 and \(\alpha=16\). Training uses a global batch size of 32 across four NVIDIA A40 GPUs, a 4,096-token context, at most 262,144 image pixels, FP16 precision, and a cosine schedule with 10\% warm-up. The learning rate is \(5\times10^{-5}\) for Box Pickup and \(1\times10^{-4}\) for Cable Threading.


\subsection{Simulation Results}

\subsubsection{{Quantitative Results}}

We use the same baselines and evaluation metrics for all three simulated tasks. For each VLA backbone, \(\pi_0\)~\cite{black2024pi_0} and \(\pi_{0.5}\)~\cite{intelligence2025pi05visionlanguageactionmodelopenworld}, we compare the base policy trained without data augmentation or additional annotations, PEEK~\cite{zhang2025peek}, and our $h^v$-guided visual prompting. Table~\ref{tab:simulation_results} reports task success rates for each method and backbone.

On Rope Pulling, our method performs comparably to the base policies and exceeds PEEK~\cite{zhang2025peek} by 12--14\% across backbones, demonstrating compatibility with tasks with no signature correction. PEEK's lower performance is likely because Rope Pulling repeatedly alternates arms that grasp and pull the rope, making it hard to decompose the end effectors' interleaved motions.

On Cable Threading, our method exceeds the base policies by 22--30\% and PEEK~\cite{zhang2025peek} by 57--58\% across the two backbones. These gains support the proposed pipeline for execution requiring a transition to \(h^{v \star}=1\). As in Rope Pulling, PEEK again struggles to distinguish the two arms during alternating motions. Here, execution must additionally respect the relationship between the robot--object and obstacle loops, making correct motion selection essential.

On Box Pickup, our method improves task exceeds the base policies by 37--53\% and PEEK~\cite{zhang2025peek} by 16--27\% across the two backbones. Together with Cable Threading, these results support the pipeline's applicability to tasks requiring different demonstrated signatures, including correction to \(h^{v \star}=0\), rather than only a single type of transition.

To isolate signature recognition from task execution, we evaluate virtual \(h^v\)-signature prediction on 100 randomly sampled unseen configurations per task. Table~\ref{tab:vlm_signature_accuracy} reports accuracy for each VLM trained on augmented data generated with its corresponding VLA backbone. Box Pickup accuracy is 16--18\% higher than Cable Threading. This is consistent with Box Pickup's higher task success and suggests that accurate signature estimation is correlated with better performance.

\begin{table}[!t]
\caption{{VLM prediction accuracy (\%) for the virtual $h^v$-signature on unseen simulation configurations.}}
\label{tab:vlm_signature_accuracy}
\centering
{
\begin{tabular}{lcc}
\hline
& \multicolumn{2}{c}{VLA backbone} \\
Task & \(\pi_0\)~\cite{black2024pi_0} & \(\pi_{0.5}\)~\cite{intelligence2025pi05visionlanguageactionmodelopenworld} \\
\hline
Cable Threading & 74 & 76 \\
Box Pickup & 92 & 92 \\
\hline
\end{tabular}
}
\end{table}

\subsubsection{{Qualitative Results}}

{Our method overlays topology-informed motion prompts on the original observation, preserving both the manipulation target and the surrounding obstacle geometries that constrain how it can be reached. The relevance of those geometries differs across tasks. Rope Pulling requires repeated grasping and pulling without modifying the topological signature. In Cable Threading and Box Pickup, however, the relationship between the robot--object loop and the obstacle loop determines whether the robot can approach the target directly or must first modify $h^v$.}

\begin{figure}[!t]
    \centering
    \includegraphics[width=1.0\linewidth]{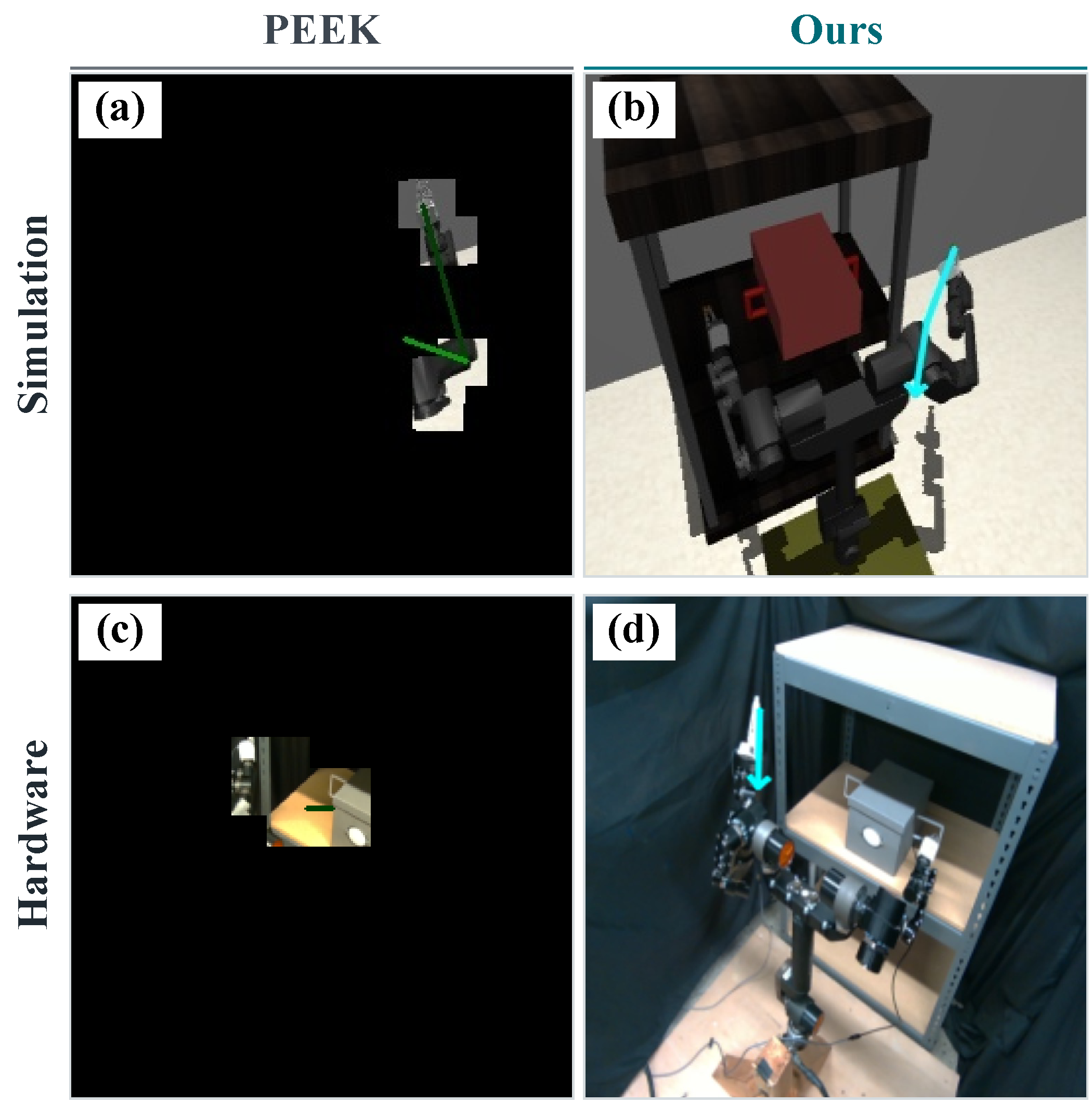}
    \caption{Visual prompting for Box Pickup in simulation with (a) PEEK and (b) ours, and on hardware with (c) PEEK and (d) ours. PEEK masks out task-relevant rack geometry, whereas our method preserves this context and provides appropriate visual prompts}
    \label{fig:prompt_comparison}
\end{figure}

The comparison with PEEK~\cite{zhang2025peek} illustrates the consequences
of excluding this scene context. PEEK~\cite{zhang2025peek} emphasizes manipulated objects and subtrajectory goals but can mask geometry required for signature reasoning. Its masks omit the obstacle loop in Cable Threading and the relevant rack posts in Box Pickup, as shown in Fig.\ref{fig:prompt_comparison}(a). Our method preserves this geometry in Fig.\ref{fig:prompt_comparison}(b). Rope Pulling has no task-relevant obstacle loop and is consequently less sensitive to this loss of context, helping explain PEEK’s smaller performance drop. Together, these results suggest that visual guidance should preserve structures that constrain the motion, even when they are not themselves manipulation targets.

\subsection{Real-World Results}
To run the Box Pickup task on hardware, we apply visual domain randomization to the colors and textures of scene objects and to the camera positions and viewing angles. To address residual sim-to-real mismatch in reaching the box handles, we collect 15 short real-world trajectories demonstrating precise approaches to the handles for lifting the box. We also collect three $h^v$-correction trajectories to reduce the risk of overfitting to the reaching demonstrations. Both the baselines and our VLA are further fine-tuned on these 18 trajectories, whose total frame count is 3\% of that in the collected simulation data.

For VLM adaptation, we human-label the \(h^v\) signature in observations from 1,000 random real-world configurations. We fine-tune the VLM on these labeled observations, which constitute approximately 5\% of its total training data. During real-world execution, once the end effectors reach the designated handle regions, a low-level controller directly lifts the box.
We do not fine-tune the PEEK VLM, as doing so would require full demonstrations on hardware which we do not assume we have.
As such, our method does not rely on hardware tele-operation data which may be challenging to collect, especially for bimanual manipulation.
At the same time, we can utilize cheaper-to-collect $h^v$-annotated data.

\subsubsection{{Quantitative Results}}

We report overall task success and $h^v$-correction success, which measures transition from an initially incorrect to the demonstrated signature.  Our method achieves 60\% task success and 83\% correction success, exceeding the strongest baselines by 40 and 53 percentage points, respectively, as shown in Table~\ref{tab:hardware_results}. The correction gain demonstrates that topology-guided prompting effectively resolves signature mismatches, while the task-success gain shows that these corrections translate into substantially better task completion on hardware.

\begin{table}[!t]
\caption{Real-world Box Pickup results. Base Policy refers to the VLA backbone fine-tuned on $\mathcal{D}_{\mathrm{nominal}}$.}
\label{tab:hardware_results}
\centering
{
\begin{tabular}{lcc}
\hline
Method & \shortstack{Overall task\\success (\%)} & \shortstack{$h^v$-correction\\success (\%)} \\
\hline
Base policy \(\pi_0\)~\cite{black2024pi_0} & 20 & 30 \\
PEEK~\cite{zhang2025peek} & 3 & 20 \\
Ours & 60 & 83 \\
\hline
\end{tabular}
}
\end{table}

\subsubsection{{Qualitative Results}}
{Fig.~\ref{fig:prompt_comparison} compares the visual prompts produced by PEEK and our method. In configurations with an incorrect signature, the nominal VLA frequently reaches directly toward the handle without moving around the rack post. This behavior resembles the direct approach demonstrated from training configurations with the demonstrated signature, but leads to obstruction in these real world initial states. The VLM trained using PEEK~\cite{zhang2025peek} similarly directs the end effector toward the handle without providing guidance around the post. In contrast, our method recognizes the incorrect signature and guides the arm around the post before approaching the handle.}

We also observe recovery from incorrect signature predictions during execution. In some trials, our method initially predicts the signature incorrectly and directs the end effector toward the handle. At a subsequent observation update, it revises the prediction and redirects the arm around the post. This illustrate how repeated signature prediction and visual-prompt updates can correct an initially inappropriate motion.

\section{Conclusion}
We presented a topology-guided visual-prompting framework that combines planning in an approximate simulator with vision-based manipulation. Privileged simulator geometry supports the generation of signature-correction and task-continuation demonstrations for VLA fine-tuning. At deployment, a fine-tuned VLM predicts task-relevant signatures and end-effector waypoints, providing visual guidance without requiring full environment geometry. Across three simulated bimanual tasks and real-world box pickup, our method outperforms a VLM-prompting baseline and improves hardware task success by 40\% over the strongest baseline. These results support using explicit signature reasoning to distinguish configurations requiring different motions, while preserving the scene context needed for execution.

\bibliographystyle{IEEEtran}
\bibliography{references}

\end{document}